\documentclass[10pt]{scrartcl}

\usepackage{amssymb}
\usepackage{amsmath}
\usepackage{amscd}
\usepackage{amsthm}

\newtheorem{definition}{Definition}[section]
\newtheorem{theorem}[definition]{Theorem}
\newtheorem{corollary}[definition]{Corollary}
\newtheorem{proposition}[definition]{Proposition}
\newtheorem{lemma}[definition]{Lemma}

\renewcommand{\qed}{\hspace*{\fill} $\blacksquare$}

\newcommand{\commentout}[1]{}

\newcommand{\introskip}{\vspace{3ex}}

\newcommand{\R}{\mathbb{R}}                    

\newcommand{\abs}[1]{\mathop{\left\lvert #1 \right\rvert}} 
\newcommand{\args}[1]{\mathop{\left( #1 \right)}} 
\newcommand{\inner}[1]{\mathop{\left\langle #1 \right\rangle}}
\newcommand{\norm}[1]{\mathop{\left\lVert #1 \right\rVert}}
\newcommand{\cbrace}[1]{\mathop{\left\{ #1 \right\}}}
\newcommand{\bracket}[1]{\mathop{\left[ #1 \right]}}

\newcommand{\argsS}[2]{\mathop{\left( #1 \right)#2}} 

\newcommand{\normS}[2]{\mathop{\left\lVert #1 \right\rVert#2}}

\renewcommand{\S}[1]{{\mathcal{#1}}}           	
\def\vec#1{\mathchoice{\mbox{\boldmath$\displaystyle#1$}}
{\mbox{\boldmath$\textstyle#1$}}
{\mbox{\boldmath$\scriptstyle#1$}}
{\mbox{\boldmath$\scriptscriptstyle#1$}}}

\renewenvironment{cases}{%
\left\{\begin{array}{c@{\quad : \quad}l}}%
{%
\end{array}\right.}

\newcounter{part_counter}
\renewenvironment{part}{
\refstepcounter{part_counter}

\medskip

\noindent
\textbf{\arabic{part_counter}.} 
}%

\usepackage{framed}

\begin{document}

\title{Asymptotic Behavior of Mean Partitions in Consensus Clustering}
\author{Brijnesh J.~Jain \\
 Technische Universit\"at Berlin, Germany\\
 e-mail: brijnesh.jain@gmail.com}
 
\date{}
\maketitle

\begin{abstract} 
Although consistency is a minimum requirement of any estimator, little is known about consistency of the mean partition approach in consensus clustering. This contribution studies the asymptotic behavior of mean partitions. We show that under normal assumptions, the mean partition approach is consistent and asymptotic normal. To derive both results, we represent partitions as points of some geometric space, called orbit space. Then we draw on results from the theory of Fr\'echet means and stochastic programming. The asymptotic properties hold for continuous extensions of standard cluster criteria (indices). The results justify consensus clustering using finite but sufficiently large sample sizes. Furthermore, the orbit space framework provides a mathematical foundation for studying further statistical, geometrical, and analytical properties of sets of partitions.  
\end{abstract} 

\newpage

\tableofcontents

\newpage

\section{Introduction}

Clustering is a standard technique for exploratory data analysis that finds applications across different disciplines such as computer science, biology, marketing, and social science. The goal of clustering is to group a set of unlabeled data points into several clusters based on some notion of dissimilarity.  Inspired by the success of classifier ensembles, consensus clustering has emerged as a research topic \cite{Ghaemi2009,VegaPons2011}. Consensus clustering first generates several partitions of the same dataset. Then it combines the sample partitions to a single consensus partition. The assumption is that a consensus partition better fits to the hidden structure in the data than individual partitions. 

One standard approach of consensus clustering combines the sample partitions to a mean partition  \cite{Dimitriadou2002,Domeniconi2009,Filkov2004,Franek2014,Gionis2007,Li2007,Strehl2002,Topchy2005,VegaPons2010}. A mean partition best summarizes the sample partitions with respect to some (dis)similarity function. A natural question is the choice of sample size. If the sequence of mean partitions fails to converge stochastically for growing sample size $n$, then picking a reasonable value for $n$ becomes an additional parameter selection problem. Otherwise, if the mean partitions converge stochastically to an expected partition, the problem of selecting a sample size $n$ simplifies to the problem of selecting a sufficiently large $n$, because we have high confidence that nothing unexpected will happen when sampling further partitions. In other words, stochastic convergence justifies the common practice to draw finite but sufficiently large sample sizes. 

Though there is an extensive literature on consensus clustering \cite{VegaPons2011}, little is known about the asymptotic behavior of the mean partition approach. Topchy et al.~\cite{Topchy2004} studied the asymptotic behavior of the mean partition approach under the following simplifying assumptions: 
\begin{enumerate}
\itemsep0em
\item[(A1)] The underlying distance is a semi-metric.\footnote{A semi-metric satisfies all axioms of a metric, but not necessarily the triangle inequality.}
\item[(A2)] Partitions are hard (crisp) partitions.
\item[(A3)] The expected partition is unique.
\item[(A4)] Partitions strongly concentrate on the expected partition. 
\end{enumerate}
In this contribution, we study the asymptotic behavior of the mean partition approach without drawing on assumptions (A2)--(A4). We show (i) consistency of the mean partitions, (ii) strong consistency of the variations, and (iii) a modified version of the Central Limit Theorem for mean partitions. We present two variants of results (i) and (ii). The first variant assumes that partitions form a compact metric space. The second variant requires the Euclidean space as ambient space and assumes that partitions are compared by a continuous cluster criterium. We also draw on continuity of the cluster criterium for showing result (iii). Since standard  criteria for comparing partitions are defined on the discrete space of hard partitions, we present examples of their continuous extensions. We can apply the generalized standard criteria to soft partitions and we can analyze the asymptotic behavior of the mean partition approach for hard and soft partitions in a unified manner. 

The basic idea to derive the results is to represent partitions as points of a geometric space, called orbit space. Orbit spaces are well explored, possess a rich mathematical structure and have a natural connection to Euclidean spaces \cite{Bredon1972,Jain2015,Ratcliffe2006}. For the first variant of results (i) and (ii), we link the consensus function of the mean partition approach to Fr\'echet functions \cite{Frechet1948}, which are well explored in mathematical statistics \cite{Bhattacharya2003,Bhattacharya2012}. The second variant of results (i) and (ii) as well as result (iii) apply results from stochastic programming \cite{Shapiro2009}.  

The rest of this paper is structured as follows: Section \ref{sec:geometry} constructs the orbit space of partitions and introduces metric structures. In Section \ref{sec:Theory}, we introduce Fr\'echet consensus functions and study their asymptotic behavior. Section \ref{sec:criteria} presents examples of continuous extensions of standard cluster criteria. Finally, Section \ref{sec:conclusion} concludes with a summary of the main results and with an outlook to further research. We present proofs in the appendix.

\section{Geometry of Partition Spaces}\label{sec:geometry}

In this section, we show that a partition can be represented as a point in some geometric space, called orbit space. Then we endow orbit spaces $\S{P}$ with metrics $\delta$ derived from the Euclidean space and study their properties.

\subsection{Partitions}
Let $\S{Z}= \cbrace{z_1, \ldots, z_m}$ be a set of $m$ data points. A partition $X$ of $\S{Z}$ with $\ell$ clusters $\S{C}_1, \ldots, \S{C}_{\ell}$ is specified by a matrix $\vec{X} \in [0,1]^{\ell \times m}$ such that $\vec{X}^T\vec{1}_\ell = \vec{1}_m$, where $\vec{1}_\ell \in \R^\ell$ and $\vec{1}_m \in \R^m$ are vectors of all ones. 

The rows $\vec{x}_{k:}$ of matrix $\vec{X}$ refer to the clusters $\S{C}_k$ of partition $X$. The columns $\vec{x}_{:j}$ of $\vec{X}$ refer to the data points $z_j \in \S{Z}$. The elements $x_{kj}$ of matrix $\vec{X} = (x_{kj})$ represent the degree of membership of data point $z_j$ to cluster $\S{C}_k$. The constraint $\vec{X}^T\vec{1}_\ell = \vec{1}_m$ demands that the membership values $\vec{x}_{:j}$ of data point $z_j$ across all clusters must sum to one.

By $\S{P}_{\ell,m}$ we denote the set of all partitions with $\ell$ clusters over $m$ data points. Since some clusters may be empty, the set $\S{P}_{\ell,m}$ also contains partitions with less than $\ell$ clusters. Thus, we consider $\ell \leq m$ as the maximum number of clusters we encounter. If the exact numbers $\ell$ and $m$ do not matter or are clear from the context, we also write $\S{P}$ for $\S{P}_{\ell, m}$. A hard partition $X$ is a partition with matrix representation $\vec{X} \in \cbrace{0,1}^{\ell \times m}$. The set $\S{P}^+ \subset \S{P}$ denotes the subset of all hard partitions.

\subsection{The Orbit Space of Partitions}

The representation space $\S{X}$ of the set $\S{P} = \S{P}_{\ell, m}$ of partitions is a set of the form
\[
\S{X} = \cbrace{\vec{X} \in [0,1]^{\ell \times m} \,:\, \vec{X}^T\vec{1}_\ell = \vec{1}_m}.
\]
Then we have a natural projection
\[
\pi: \S{X} \rightarrow \S{P}, \quad \vec{X} \mapsto X = \pi(\vec{X})
\]
that sends matrices $\vec{X}$ to partitions $X$ they represent. The map $\pi$ conveys two properties: (1) each partition can be represented by at least one matrix, and (2) a partition may have several matrix representations. 
 
Suppose that matrix $\vec{X} \in \S{X}$ represents a partition $X \in \S{P}$. The subset of all matrices representing $X$ forms an equivalence class $\bracket{\vec{X}}$ that can be obtained by permuting the rows of matrix $\vec{X}$ in all possible ways. The equivalence class of $\vec{X}$, called orbit henceforth, is of the form
\[
\bracket{\vec{X}} = \cbrace{\vec{PX} \,:\, \vec{P} \in \Pi},
\]
where $\Pi$ is the group of all ($\ell \times \ell$)-permutation matrices. The orbit space of partitions is the set 
\[
\S{X}/\Pi = \cbrace{\bracket{\vec{X}} \,:\, \vec{X} \in \S{X}}. 
\]
The orbit space consists of all orbits $\bracket{\vec{X}}$, we can construct as described above. Mathematically, the orbit space $\S{X}/\Pi$ is the quotient space obtained by the action of the permutation group $\Pi$ on the set $\S{X}$. The orbits $[\vec{X}]$ are in 1-1-correspondence with the partitions $X = \pi(\vec{X})$. Therefore, we can identify partitions with orbits and occasionally write $\vec{X} \in X$ if $X = \pi(\vec{X})$. 

\subsection{Metric Structures}

This section endows the partition space $\S{P}$ with metrics that are derived by a generic construction principle. As examples, we consider metrics $\delta_p$ derived from  $l_p$-metrics of Euclidean spaces. We show that $\args{\S{P}, \delta_p}$ is a compact metric space for $p \geq 1$, and $\args{\S{P}, \delta_2}$ is a geodesic space.

\introskip

Every metric $d$ on the representation space $\S{X} \subset \R^{\ell \times m}$ induces a distance function
\begin{align}\label{eq:def_orbit_distance}
\delta: \S{P} \times \S{P} \rightarrow \R, \quad (X, Y) \mapsto \min \cbrace{d\!\args{\vec{X}, \vec{Y}}\,:\, \vec{X} \in X, \vec{Y} \in Y}.
\end{align}
Note that the minimum in \eqref{eq:def_orbit_distance} exists, because the orbits $\bracket{\vec{X}}$ and $\bracket{\vec{Y}}$ are finite. As an example, we consider distance functions induced by the $l_p$-norm. The $l_p$-norm for matrices $\vec{X} \in \S{X}$ is defined by
\[
\normS{\vec{X}}{_p} = \argsS{\sum_{k = 1}^\ell \sum_{j = 1}^m \abs{x_{kj}}{^p}}{^{1/p}}
\]
for every $p \geq 1$. The $l_p$-norm induces the distance function
\[
\delta_p: \S{P} \times \S{P} \rightarrow \R, \quad (X, Y) \mapsto \min \cbrace{\normS{\vec{X} - \vec{Y}}{_p}\,:\, \vec{X} \in X, \vec{Y} \in Y},
\]
called $l_p$-distance on $\S{P}$, henceforth.

To show that distances $\delta$ on $\S{P}$ induced by metrics $d$ on $\S{X}$ are also metrics, we demand that metric $d$ is permutation invariant. We say, a metric $d$ on $\S{X}$ is permutation invariant, if 
\[
d(\vec{PX}, \vec{PY}) = d(\vec{X}, \vec{Y}) 
\]
for all permutations $\vec{P} \in \Pi$. Permutation invariance means that the metric $d$ is invariant under simultaneously relabeling the clusters of  $\vec{X}$ and $\vec{Y}$. An example of permutation invariant metrics are the $l_p$-metrics. The next result shows that permutation invariant metrics on $\S{X}$ induce distances on $\S{P}$ that are again metrics. 

\begin{theorem}\label{theorem:geodesic-space}
Let $\args{\S{X}, d}$ be a metric space and let $(\S{P}, \delta)$ be the partition space endowed with distance function $\delta$ induced by metric $d$. Suppose that $d$ is permutation invariant. Then we have:
\begin{enumerate}
\item The distance $\delta$ is a metric.
\item $\args{\S{P}, \delta}$ is a compact space. 
\item $\args{\S{P}, \delta_2}$ is a geodesic space.
\end{enumerate}
\end{theorem}

Theorem \ref{theorem:geodesic-space} presents a generic way to construct metrics on $\S{P}$. Being a compact metric space is a strong property for consistency statements. Being a geodesic space means that any pair of partitions $X$ and $Y$ have a midpoint partition $M$ such that 
\[
\delta_2(X, M) = \delta_2(Y, M) = \frac{1}{2}\delta_2(X, Y).
\]
Note that being a geodesic space is a necessary and sufficient condition for guaranteeing the midpoint property for all pairs of partitions.

\section{ Fr\'echet Consensus Clustering}\label{sec:Theory}

This section first formalizes the problem of consensus clustering using the mean partition approach and then studies its asymptotic behavior. For this, we link the consensus function of the mean partition approach to the Fr\'echet function \cite{Frechet1948} from mathematical statistics. Then we show that under normal conditions the mean partition approach is consistent and asymptotically normal.

\subsection{Fr\'{e}chet Functions}

Let $\rho: \S{P} \times \S{P} \rightarrow \R$ be a (dis)similarity function on the partition space $\S{P}$. To avoid case distinctions between min- and max-operations, we assume that $\rho = -s$ if $s$ is a similarity function. Typically, we may think of $\rho$ as a criterium for comparing partitions such as, for example, the metrics $\delta_p$, the Mirkin distance, the Rand index, and the variation of information \cite{Meila2007}.  

We assume that $Q$ is a probability distribution on the partition space $\S{P}$. Suppose that $\S{S}_n = \args{X_1, X_2, \ldots, X_n}$ is a sample of $n$ partitions $X_i \in \S{P}$ drawn i.i.d.~from the probability distribution $Q$. Then 
\begin{align}\label{eq:sff}
F_n: \S{P} \rightarrow \R, \quad Z \mapsto \frac{1}{n}\sum_{i=1}^n \rho\!\args{X_i, Z}
\end{align}
is the Fr\'{e}chet function of $\S{S}_n$ corresponding to the dissimilarity function $\rho$. The infimum $V_n$ of $F_n$ is the variation of $\S{S}_n$. The minimizers of the Fr\'echet function need neither exist nor be unique. The (possibly empty) set
\[
\S{F}_n = \cbrace{M \in \S{P} \,:\, M = \arg\min_Z F_n(Z)}
\]
is called the mean partition set of $\S{S}_n$. Every $M \in \S{F}_n$ is a mean partition of $\S{S}_n$. 

In statistics, the mean is an estimator of the expectation. To adopt this idea to partitions, we define the expected Fr\'{e}chet function of $Q$ corresponding to the dissimilarity function $\rho$ as
\[
F_Q: \S{P} \rightarrow \R, \quad Z \mapsto \int_{\S{P}} \rho(X, Z)\, dQ(X).
\]
The infimum $V_Q$ of $F_Q$ is the expected variation of $Q$. The set 
\[
\S{F}_Q = \cbrace{M \in \S{P} \,:\, M = \arg\min_Z F_Q(Z)}
\]
is called the expected partition set of $Q$. Any $M \in \S{F}_Q$ is an expected partition of $Q$.

\subsection{Consistency by Metric Structures}
This section shows consistency of the mean partition set $\S{F}_n$ and the variation $V_n$ under the assumption that the dissimilarity function $\rho$ is of the form
\[
\rho = h \circ \delta,
\]
where $\delta$ is a metric on $\S{P}$ and $h: \R_+ \rightarrow \R_+$ is a continuous non-negative loss function. The consistency result of this section is the first of two variants. The distinguishing feature of this variant is that consistency of mean partitions is shown in an abstract metric space without connection to an ambient Euclidean space. 
 
\introskip

Roughly, we may think of the mean partition set $\S{F}_n$ being a strongly consistent estimator of the expected partition set $\S{F}_Q$, if
\[
P\args{\lim_{n \to \infty} \S{F}_n \subseteq \S{F}_Q} = 1.
\]
A precise definition of strong consistency in the sense of Bhattacharya-Patrangenaru (BP) is given in Section \ref{sec:BP-consistency}. We have the following existence and consistency result:

\begin{theorem}\label{theorem:consistency}
Suppose that $\args{\S{P}, \delta}$ is a compact metric space and $h:\R_+ \rightarrow \R_+$ is a continuous loss function. Then the following holds:
\begin{enumerate}
\item 
$\S{F}_Q$ are non-empty and compact.
\item 
$\S{F}_n$ is a BP-strongly consistent estimator of $\S{F}_Q$.
\item 
If $\S{F_Q} = \cbrace{M}$, then $\S{F}_n$ is a strongly consistent estimator of $M$.
\item 
$V_n$ is a strongly consistent estimator of $V_Q$.
\end{enumerate}
\end{theorem}

\noindent
\proof
The assertions of Theorem \ref{theorem:consistency} follow from \cite{Bhattacharya2012}, Theorem 2.3, Prop.~2.8, and Corollary 2.4. 
\qed

\medskip

The role of the loss function $h$ is to generalizes the concept of Fr\'echet function of order $p$ corresponding to $\rho(X, Y) = \delta(X, Y)^p$. In this case, the loss is of the form $h(x)= x^p$. For $p=1$, we recover the consensus function of the median partition  and for $p= 2$ the consensus function of the mean partition. 

Immediate consequences of Theorem \ref{theorem:geodesic-space} and \ref{theorem:consistency} are as follows:
\begin{corollary}
Let $(\S{P}, \delta)$ be a partition space. Assertions (1)--(4) of Theorem \ref{theorem:consistency} hold for the following special cases:
\begin{enumerate}
\item $\delta$ is induced by a permutation invariant metric on $\S{X}$.
\item $\delta = \delta_p$ is an $l_p$-metric on $\S{P}$.
\end{enumerate} 
\end{corollary}

\subsection{Consistency by Stochastic Programming}\label{subsec:constop}

In this section, we consider the following stochastic programming problem:
\begin{align}\label{eq:stop}
\min_{Z \in \S{P}} \cbrace{ F_Q(Z) = \int_{\S{P}} \rho(X, Z)\, dQ(X)},
\end{align}
where $\rho(X, Z)$ is a continuous dissimilarity function and $X$ is a random partition with probability distribution $Q$. The consistency result of this section is the second of both variants. This variant requires no metric structure of the dissimilarity function. However, the proof compensates the missing structure of $\rho$ by exploiting the structure of the ambient Euclidean space.

\introskip

To establish consistency, we need to introduce two concepts: (1) continuity of a function on partitions, and (2) distance between partitions. 

To define both concepts, we assume the geodesic space $\args{\S{P}, \delta_2}$, where $\delta_2$ is the $l_2$-metric induced by the Euclidean norm. We say, a function $f:\S{P} \rightarrow \R$ is continuous at partition $Z$, if for every $\varepsilon > 0$ there is a $\zeta > 0$ such that for all $X \in \S{P}$ with $\delta_2(X, Z) < \zeta$ the value $f(X)$ satisfies $\abs{f(X)-f(Z)} < \varepsilon$. Furthermore, by 
\[
D(\S{U}, \S{V}) = \sup_{X \in \,\S{U}} \inf_{\phantom{^X}Y \in \S{V}\phantom{^X}} \delta_{2}(X, Y)
\]
we denote the distance between two subsets $\S{U}, \S{V} \subseteq \S{P}$. 

\medskip

The next theorem gives convergence results based on the continuity of the dissimilarity function $\rho$.
\begin{theorem}\label{theorem:constop}
Let $X_1, \ldots, X_n \in \S{P}$ be a sample of $n$ partitions drawn i.i.d.~from the probability distribution $Q$. Suppose that $\rho(X, Z)$ is continuous in both arguments. Then
\begin{enumerate}
\item $D\!\args{\S{F}_n, \S{F}_Q} \rightarrow 0$ almost surely.
\item $V_n \rightarrow V_Q$ almost surely.
\end{enumerate}
\end{theorem}

Section \ref{sec:criteria} presents examples of continuous dissimilarity functions.The examples are continuos extensions of standard criteria for comparing partitions. For all these examples, the mean partition approach is consistent in the sense of Theorem \ref{theorem:constop} under mild assumptions.

\subsection{A Central Limit Theorem}

Consistency of the mean partition approach gives a certain assurance that the error of the estimation tends to zero in the limit as the sample size grows to infinity. It does, however, not indicate the magnitude of the error for a given sample size. In standard statistics, we can use the Central Limit Theorem (CLT) to construct an approximate confidence interval for the unknown population mean and derive the order of the estimation error as a function of the sample size. To lay the foundations for these techniques, we present a variant of the CLT for mean partitions.
 
\introskip

We assume the same setting as in Section \ref{subsec:constop}. By $\S{N}(0, \sigma^2)$ we denote the normal distribution with mean zero and variance $\sigma^2$. Convergence of random variable $(X_n)$ to a random variable $X$ in distribution is denoted by $X_n \xrightarrow{\;d\;} X$. The next result states a version of the CLT for mean partitions. 

\begin{theorem}\label{theorem:CLT}
Suppose that every sample $\S{S}_n$ of $n$ partitions is drawn i.i.d.~from $\S{P}$. Then 
\[
\frac{V_n - V}{\sqrt{n}} \;\xrightarrow{\;\;d\;\;}\; \inf_{M \in \S{F}_Q}\S{N}(0,\sigma^2(M)).
\]
\end{theorem}

\medskip

If the expected partition set $\S{F}_Q = \cbrace{M}$ consists of a singleton $M$, then Theorem \ref{theorem:CLT} reduces to 
\[
\frac{V_n - V}{\sqrt{n}} \;\xrightarrow{\;\;d\;\;}\; \S{N}(0,\sigma^2(M)).
\]

\section{Generalized Criteria for Comparing Partitions}\label{sec:criteria}

This section extends standard criteria for comparing hard partitions to criteria for comparing arbitrary partitions. There are infinitely many ways to extend functions from a discrete to a continuous domain. We suggest extensions that partially admit a probabilistic interpretation. Continuity of a criterium allows us to invoke Theorem \ref{theorem:constop} and \ref{theorem:CLT}. For details on criteria based on counting pairs and cluster matchings, we refer to \cite{Meila2007} and for details on information-theoretic criteria, we refer to \cite{Vinh2010}.

\subsection{Criteria based on Counting Pairs}

A hard partition $X \in \S{P}^+$ induces an equivalence relation 
\[
z \sim_X z' \quad \Leftrightarrow \quad z \text { and } z' \text{ are in the same cluster of } X
\]
for all $z, z' \in \S{Z}$. Let $\S{Z}^{\bracket{2}}$ denote the set of all 2-element subsets of $\S{Z}$. The confusion matrix $\vec{C}^+(X,Y) = (m_{pq}^+)$ of hard partitions $X$ and $Y$ is a ($2 \times 2$)-matrix with elements
\begin{align*}
m_{11}^+ &= \abs{\cbrace{\cbrace{z, z'} \in \S{Z}^{[2]} \,:\, z \sim_X z', z \sim_Y z'}}\\
m_{10}^+ &= \abs{\cbrace{\cbrace{z, z'} \in \S{Z}^{[2]} \,:\, z \sim_X z', z \nsim_Y z'}}\\
m_{01}^+ &= \abs{\cbrace{\cbrace{z, z'} \in \S{Z}^{[2]} \,:\, z \nsim_X z', z \sim_Y z'}}\\
m_{00}^+ &= \abs{\cbrace{\cbrace{z, z'} \in \S{Z}^{[2]} \,:\, z \nsim_X z', z \nsim_Y z'}}.
\end{align*}
The matrix $\vec{C}^+(X,Y)$ satisfies 
\[
\normS{\vec{C}^+(X,Y)}{_1} = m_{11}^+ +m_{10}^+ + m_{01}^+ + m_{00}^+ = \frac{m(m-1)}{2} = \abs{\S{Z}^{[2]}}.
\]
By $N = m(m-1)/2$ we denote the cardinality of the set $\S{Z}^{[2]}$. Criteria based on counting pairs can be described by a function of the general form
\[
\rho(X, Y) = f\Big(\vec{C}^+(X, Y)\Big), 
\]
where $X, Y \in \S{P}^+$. To extend $\rho$ to the space $\S{P}$ of all partitions, we define the compatibility matrix of a partition $X \in \S{P}$ as an ($m \times m$)-matrix of the form
\[
\vec{C}_{\!X} = \vec{X}^T\!\vec{X},
\]
where $\vec{X}$ is an arbitrary representation of $X$. From Prop.~\ref{prop:Cx-is-well-defined} follows that $\vec{C}_{\!X}$ is independent from the particular choice of representation $\vec{X} \in X$. The compatibility matrix $\vec{C}_{\!X} = \args{c_{rs}}$ consists of elements of the form
\[
c_{rs} = \inner{\vec{x}_{:r}, \vec{x}_{:s}}
\]
for all $r, s \in \cbrace{1, \ldots, m}$. Thus, $c_{rs}$ is the inner product of columns $\vec{x}_{:r}$ and $\vec{x}_{:s}$ of matrix $\vec{X}$. Recall that a column $\vec{x}_{:j}$ of a representation matrix $\vec{X}$ summarizes the membership values of data point $z_j \in \S{Z}$. 

Next, we define the function 
\[
\chi(\vec{A}, \vec{B}) = \sum_{r=1}^m \sum_{s= r+1}^m a_{rs}b_{rs}
\]
for all ($m\times m$)-matrices $\vec{A} = (a_{rs})$ and $\vec{B} = (b_{rs})$. The function $\chi(\vec{A}, \vec{B})$ is the inner product of the strictly upper triangular matrices of $\vec{A}$ and $\vec{B}$. By $\vec{1} = \vec{1}_{m \times m}$ we denote the ($m \times m$)-matrix of all ones. 

Now we can extend the confusion matrix $\vec{C}^+(X,Y)$ to its continuous counterpart $\vec{C}(X,Y) = (m_{pq})$ with elements
\begin{align*}
m_{11} &= \chi\args{\vec{C}_{\!X}, \vec{C}_{\!Y}}\\
m_{10} &= \chi\args{\vec{C}_{\!X}, \vec{1} - \vec{C}_{\!Y}}\\
m_{01} &= \chi\args{\vec{1} - \vec{C}_{\!X}, \vec{C}_{\!Y}}\\
m_{00} &= \chi\args{\vec{1} - \vec{C}_{\!X}, \vec{1} - \vec{C}_{\!Y}},
\end{align*}
The next result shows that matrix $\vec{C}(X,Y)$ indeed generalizes the confusion matrix $\vec{C}^+(X,Y)$ and admits a probabilistic interpretation.
\begin{proposition}\label{prop:properties-of-C(X,Y)}
Let $X, Y \in \S{P}$ be partitions of dataset $\S{Z}$ consisting of $m$ elements. The matrix $\vec{C}(X, Y) = (m_{pq})$ satisfies the following properties:
\begin{enumerate}
\item If $X, Y \in \S{P}^+$, then $\vec{C}^+(X, Y) = \vec{C}(X, Y)$.
\item $m_{pq} \geq 0$ for all $p,q \in \cbrace{0,1}$.
\item $\sum_{p,q} m_{pq} = N$.
\end{enumerate}
\end{proposition}
The first assertion of Prop.~\ref{prop:properties-of-C(X,Y)} states that $\vec{C}(X,Y)$ is an extension of the confusion matrix $\vec{C}^+(X,Y)$. The second and third assertion admit a probabilistic interpretation of the values $m_{pq}$ as follows:
\begin{align*}
\mathbb{P}(z \sim_X z', z\sim_Y z') &= \frac{m_{11}}{\abs{\S{Z}^{[2]}}}\\
\mathbb{P}(z \sim_X z', z\nsim_Y z') &= \frac{m_{10}}{\abs{\S{Z}^{[2]}}}\\
\mathbb{P}(z \nsim_X z', z\sim_Y z') &= \frac{m_{01}}{\abs{\S{Z}^{[2]}}}\\
\mathbb{P}(z \nsim_X z', z\nsim_Y z') &= \frac{m_{00}}{\abs{\S{Z}^{[2]}}},
\end{align*}
where $\cbrace{z, z'} \in \S{Z}^{[2]}$. The probabilistic interpretation of $m_{pq}$ is consistent with the probabilistic interpretation of $m_{pq}^+$ for hard partitions. 

\medskip

We can extend criteria $\rho(X, Y) = f(\vec{C}^+(X, Y))$ based on counting pairs from the space $\S{P}^+$ of hard partitions to the space $\S{P}$ of all partitions by replacing the confusion matrix $\vec{C}^+(X, Y)$ with matrix $\vec{C}(X, Y)$. We present extensions of some common examples:
\begin{align*}
\rho_1(X, Y) &= \frac{m_{11}}{m_{11} + m_{01}} \tag*{\text{(Wallace I)}}\\[1ex]
\rho_2(X, Y) &= \frac{m_{11}}{m_{11} + m_{01}} \tag*{\text{(Wallace II)}}\\[1ex]
\rho_3(X, Y) &= \frac{m_{11} + m_{00}}{N} \tag*{\text{(Rand)}}\\[1ex]
\rho_4(X, Y) &= \frac{m_{11}}{\sqrt{\args{m_{11} + m_{10}}\args{m_{11} + m_{01}}}} \tag*{\text{(Fowlkes-Mallows)}}\\[1ex]
\rho_5(X, Y) &= \frac{m_{11}}{m_{11} + m_{01} + m_{01}}\tag*{\text{(Jacard)}}
\end{align*}

\medskip
The next result shows that criterium $\rho_3$ is continuous and criteria $\rho_1$ -- $\rho_5$ are continuous almost everywhere. 

\begin{proposition}\label{prop:continutity_1-5}
Suppose that $m > 1$. Then criteria $\rho_1 - \rho_5$ are continuous in both arguments almost everywhere. Criterium $\rho_3$ is continuous. 
\end{proposition}

Critical points occur when we admit partitions for which $m_{11} = 0$. This situation refers to the degenerated case where each cluster contains at most one data point with positive membership value. The simplest way to cope with this issue is to consider closed subsets $\S{P}'$ of $\S{P}$ that exclude degenerated partitions.  Then Theorem \ref{theorem:constop} and \ref{theorem:CLT} can be applied to $\rho_1$ -- $\rho_5$ restricted to the compact subset $\S{P}'$.

\subsection{Criteria based on Cluster Matchings}
Every hard partition $X$ defines $\ell$ clusters $\S{C}_1(X), \ldots, \S{C}_\ell(X)$ that partition the data set $\S{Z}$. For two hard partitions $X$ and $Y$, we define the numbers
\begin{align*}
m &= \abs{\S{Z}}\\
x_p^+ &= \abs{\S{C}_p(X)}\\
y_q^+ &= \abs{\S{C}_q(Y)}\\
z_{pq}^+ &= \abs{\S{C}_p(X) \cap \S{C}_q(Y)}
\end{align*}
for all $p, q \in \cbrace{1, \ldots, \ell}$. Criteria based on cluster matchings can be described by a function of the general form
\[
\rho(X, Y) = f\Big(m, \args{x_p^+}, \args{y_q^+}, \args{z_{pq}^+}\Big), 
\]
where $X, Y \in \S{P}^+$. Let $\vec{X}$ and $\vec{Y}$ be representations of partitions $X$ and $Y$ from $\S{P}$. To extend a criterium $\rho$ to the space of all partitions, we define the values
\begin{align*}
x_p &= \inner{\vec{x}_{p:}, \vec{1}_m}\\
y_q &= \inner{\vec{y}_{q:}, \vec{1}_m}\\
z_{pq} &= \inner{\vec{x}_{p:}, \vec{y}_{q:}}
\end{align*}
for all $p, q \in \cbrace{1, \ldots, \ell}$. Recall that the rows $\vec{x}_{p:}$ and $\vec{y}_{q:}$ of $\vec{X}$ and $\vec{Y}$represent the clusters $\S{C}_p(X)$ and $\S{C}_q(Y)$. 

The next result shows that $(x_p)$, $(y_p)$, $(z_{pq})$ generalize $(x_p^+)$, $(y_p^+)$, $(z_{pq}^+)$  and partially admit a probabilistic interpretation.
\begin{proposition}\label{prop:properties-of-xyz}
Let $X, Y \in \S{P}$ be partitions of dataset $\S{Z}$ consisting of $m$ elements. Then we have:
\begin{enumerate}
\item If $X, Y \in \S{P}^+$, then $x_p^+ = x_p$, $y_q^+ = y_q$, and $z_{pq}^+ = z_{pq}$.
\item $x_{p}, y_q, z_{pq} \geq 0$ for all $p,q \in \cbrace{1, \ldots, \ell}$.
\item $\sum_p x_p = \sum_q y_q = m$
\end{enumerate}
\end{proposition}

\medskip

The first assertion of Prop.~\ref{prop:properties-of-C(X,Y)} states that $(x_p)$, $(y_p)$, $(z_{pq})$  are extensions. The second and third assertion admit a probabilistic interpretation as follows:
\begin{align*}
\mathbb{P}(z \in \S{C}_p(X)) &= \frac{x_p}{\abs{\S{Z}}}\\
\mathbb{P}(z \in \S{C}_q(Y)) &= \frac{y_q}{\abs{\S{Z}}}
\end{align*}
Next, we extend criteria $\rho(X, Y) = f\args{m, \args{x_p^+}, \args{y_q^+}, \args{z_{pq}^+}}$ based on cluster matchings from the space $\S{P}^+$ of hard partitions to the space $\S{P}$:
\begin{align*}
\rho_6(X, Y) &= \sum_p x_p^2 + \sum_q y_q^2 - \sum_{p} \sum_q z_{pq}^2 \tag*{\text{(Mirkin)}}\\[1ex]
\rho_7(X, Y) &= \max_{\vec{\phi} \in \text{Sym}_\ell} \; \frac{1}{m} \sum_{p} z_{p\phi(q)} \tag*{\text{(Meil\u{a}-Heckerman)}}\\[1ex]
\rho_8(X, Y) &= 2m - \sum_{p} \max_q z_{pq} - \sum_{q} \max_p z_{pq} \tag*{\text{(van Dongen)}}
\end{align*}
The set $\text{Sym}_\ell$ in the Meil\u{a}-Heckerman formula is the set of all permutations of the numbers $1, \ldots, \ell$.

Since criteria $\rho_6$--$\rho_8$ are continuous, the mean partitions are consistent estimators of the expected partitions according to Theorem \ref{theorem:constop} and asymptotic normal according to Theorem \ref{theorem:CLT}. 

\subsection{Criteria based on Information Theory}

Suppose that $X$ and $Y$ are hard partitions. Similarly as criteria based on cluster matchings, criteria based on Information Theory can be described by a function of the general form
\[
\rho(X, Y) = f\Big(m, \args{x_p^+}, \args{y_q^+}, \args{z_{pq}^+}\Big), 
\]
where $X, Y \in \S{P}^+$. In contrast to criteria based on cluster matchings, the function $f$ is composed of information-theoretic measures. Using the same notation as in the previous sections, continuous extensions of information-theoretic criteria for partitions $X, Y \in \S{P}$ are based on the following measures:
\begin{align*}
H(X) &= - \sum_p \frac{x_p}{N} \log  \frac{x_p}{N} \tag*{(entropy)}\\
H(X, Y) &= -\sum_p \sum_q \frac{z_{pq}}{N} \log \frac{z_{pq}}{N} \tag*{(joint entropy)}\\
H(X|Y) &= -\sum_p \sum_q \frac{z_{pq}}{N} \log \frac{z_{pq}/N}{y_q/N} \tag*{(conditional entropy)}\\
I(X, Y) &= \sum_p\sum_q \frac{z_{pq}}{N} \log \frac{z_{pq}/N}{x_p y_q/N^2} \tag*{(mutual information)}.
\end{align*}
Recall that $N = m(m-1)/2$. Continuous extensions of information-theoretic cluster criteria take the general form
\[
\rho(X, Y) = f\Big(H(X), H(X, Y), H(X|Y), I(X, Y)\Big)
\]
for all $X, Y \in \S{P}$. For examples of information-theoretic measures, we refer to \cite{Vinh2010}.

By considering closed subsets of $\S{P}$ that exclude degenerated partitions, we can invoke Theorem \ref{theorem:constop} and \ref{theorem:CLT} for function $f$ that are continuous almost everywhere.

\section{Conclusion}\label{sec:conclusion}

Under normal assumptions, consensus clustering based on the mean partition approach is consistent and asymptotic normal. To derive these results, we represented partitions as points of an orbit space and established links to the theory of Fr\'echet means and stochastic programming. The results hold for continuous extensions of standard cluster criteria under mild assumptions. 

The orbit space approach provides a mathematical foundation for studying further geometrical, analytical, and statistical properties of sets of partitions with applications not confined to clustering but for any domain that studies equivalence relations. The theoretical results suggest to apply algorithms from stochastic optimization for minimizing Fr\'echet consensus functions.

\begin{appendix}
\small
\section{Bhattacharya-Patrangenaru Consistency}\label{sec:BP-consistency}

The metric $\delta$ of the partition space $\args{\S{P}, \delta}$ induces a Borel $\sigma$-algebra $\S{B}$ on $\S{P}$ such that $\args{\S{P}, \S{B}}$ is a measurable space. Let $\args{\Omega, \S{A}, P}$ be an abstract probability space. We assume that $\args{\Omega, \S{A}, P}$ is complete in the sense that every subset of every null-set is measurable. 

A random element $X$ taking values in $\S{P}$ is a mapping $X: \Omega \rightarrow \S{P}$ measurable with respect to the Borel $\sigma$-algebra $\S{B}$. Then $Q = P \circ X^{-1}$ is a probability measure on the measurable space $\args{\S{P}, \S{B}}$.

The sample mean set $\S{F}_n$ is a Bhattacharya-Patrangenaru (BP) strongly consistent estimator of the mean set $\S{F}_Q$, if $\S{F}_Q \neq \emptyset$ and if for every $\varepsilon > 0$ and for almost every $\omega \in \Omega$ there is an integer $N = N(\varepsilon, \omega)$ such that 
\[
\S{F}_n \subseteq \S{F}_Q^\varepsilon = \cbrace{Z \in \S{P} \,:\, \delta\args{Z, \S{F}_Q} \leq \varepsilon}
\]
for all $n \geq N$ \cite{Bhattacharya2003,Bhattacharya2012}.

\section{Proofs}

\subsection{Proof of Theorem \ref{theorem:geodesic-space}}

As preparation for the third assertion of Theorem \ref{theorem:geodesic-space}, we first show that partitions spaces $\args{\S{P}_{\ell,m}, \delta_2}$ can be isometrically embedded into graph spaces endowed with a graph edit kernel metric as defined in \cite{Jain2015}. 

\medskip

Let $\S{A} = \R^m$ be the set of node and edge attributes. An attributed graph of order $\ell$ is a triple $X = (\S{V}, \S{E}, \alpha)$, where $\S{V}$ represents a set of $\ell$ nodes, $\S{E} \subseteq \S{V} \times \S{V}$ a set of edges, and $\alpha: \S{V} \times \S{V} \rightarrow \S{A}$ is an attribute function satisfying the following properties for all $i, j \in \S{V}$ with $ i \neq j$:
\begin{enumerate}
\item $\alpha(i, j) \neq \vec{0}$ for all edges $(i, j) \in \S{E}$ 
\item $\alpha(i, j) = \vec{0}$ for all non-edges $(i, j) \notin \S{E}$. 
\end{enumerate}
As opposed to edges, node attributes $\alpha(i,i)$ may take any value from $\S{A}$. By $\S{G}_{\ell, m}$ we denote the set of all attributed graphs of order $\ell$ with attributes from $\S{A} = \R^m$. The graph edit kernel metric is a function of the from
\[
\delta_g: \S{G}_{\ell,m} \times \S{G}_\ell \rightarrow \R, \quad (X, Y) \mapsto \min_{\phi \in \text{Sym}_\ell} \,\sum_{i,j} \normS{\alpha(i, j) - \alpha(\phi(i), \phi(j))}{^2},
\]
where $\text{Sym}_\ell$ is the symmetric group of permutations that can be performed on a set of cardinality $\ell$. The pair $\args{\S{G}_{\ell,m}, \delta_g}$ is called graph edit kernel space of order $\ell$. 

\begin{theorem}\label{theorem:geks}
There is an isometric embedding of the orbit space $\args{\S{P}_{\ell, m}, \delta_2}$ into the graph edit kernel space $\args{\S{G}_{\ell,m}, \delta_g}$. 
\end{theorem}

\noindent
\proof
Let $X \in \S{P} = \S{P}_{\ell, m}$ be a partition with matrix representation $\vec{X}$. A partition $X$ can be equivalently expressed by a graph $G_X = (\S{V}, \emptyset, \alpha)$ as follows: The set $\S{V}$ consists of $\ell$ nodes, each of which represents a cluster. The set of edges is empty. The attribute function is of the form 
\[
\alpha(k, j) = \begin{cases}
\vec{x}_{k:} & k = j\\
\vec{0} & k \neq j
\end{cases},
\]
where $\vec{x}_{k:}$ is the $k$-th row of matrix $\vec{X}$ referring to the $k$-th cluster of $X$. We call $G_X$ the partition graph of $X$. This construction gives an injective map $f: \S{P}_{\ell,m} \rightarrow \S{G}_{\ell,m}$. 

Next, we show that the map $f$ is isometric. Observe that a graph $G = \args{\S{V, \S{E}}, \alpha} \in \S{G}_{\ell,m}$ can be represented by a $(\ell \times \ell)$-matrix $\vec{G} = (\vec{g}_{ij})$ with elements $\vec{g}_{ij} = \alpha(i, j)$. Then the graph edit kernel distance between graphs $G$ and $H$ with matrix representations $\vec{G}$ and $\vec{H}$, resp., can be equivalently expressed by 
\[
\delta_g(G, H) = \min_{\vec{P} \in \Pi} \normS{\vec{G} - \vec{P}\vec{H}\vec{P}^T}{_2},
\]
where $\Pi$ is the set of all $(\ell \times \ell)$-permutation matrices \cite{Jain2015}. Let $X, Y \in \S{P}_{\ell,m}$ be two partitions with respective matrix representations $\vec{X}$ and $\vec{Y}$. Suppose that $G_X$ and $G_Y$ are the partition graphs of $X$ and $Y$, resp., with respective matrix representations $\vec{G}_X$ and $\vec{G}_Y$. Since $\vec{G}_X$ and $\vec{G}_Y$ are diagonal matrices, we have
\[
\normS{\vec{X} - \vec{P}\vec{Y}}{_2} = \normS{\vec{G}_X - \vec{P}\,\vec{G}_Y\vec{P}^T}{_2}
\]
for all $\vec{P}\in \Pi$. This shows
\[
\delta_2(X, Y) = \min_{\vec{P} \in \Pi} \normS{\vec{X} - \vec{P}\vec{Y}}{_2} = \min_{\vec{P} \in \Pi} \normS{\vec{G}_X - \vec{P}\,\vec{G}_Y\vec{P}^T}{_2} = \delta_g\args{\vec{G}_X, \vec{G}_Y}.
\]
Thus, $f$ is an isometric embedding. 
\qed

\paragraph*{Proof of Theorem \ref{theorem:geodesic-space}.}  
Note that $\Pi$ is a group of isometries acting on $\S{X}$, because metric $d$ is invariant under permutations by assumption. 
\setcounter{part_counter}{0}
\begin{part}
Since $\Pi$ is finite, each orbit $\bracket{\vec{X}}$ is a finite and therefore a closed subset of $\S{X}$. Then the distance function $\delta$ induced by metric $d$ is a metric by \cite{Ratcliffe2006}, Theorem 6.6.1.
\end{part}

\begin{part} 
By construction of $\delta$, the group $\Pi$ is a finite group of isometries acting on the space $\R^{\ell \times m}$. Then all orbits $\bracket{\vec{X}}$ are finite and closed subsets of $\R^{\ell \times m}$. Since the Euclidean space is a finitely compact metric space, the quotient space $\R^{\ell \times m} /\Pi$ is a complete metric space by \cite{Ratcliffe2006}, Theorem 8.5.2. Since $\S{X} \subseteq \R^{\ell \times m}$ is compact and $\Pi$ is finite and therefore a compact group, compactness of $\S{P}$ follows from \cite{Bredon1972}, Chapter I, Theorem 3.1.
\end{part}

\begin{part}
We show that $\args{\S{P}_{\ell,m}, \delta_2}$ is a geodesic space. From \cite{Jain2015}, Theorem 3.3 follows that $\args{\S{G}_{\ell,m}, \delta_g}$ is a geodesic space. Let $f:\S{P}_{\ell,m} \rightarrow \S{G}_{\ell,m}$ be the isometric embedding defined in the proof of Theorem \ref{theorem:geks}. Suppose that $X$ and $Y$ are partitions from $\S{P}_{\ell,m}$ and $G_X = f(X)$ and $G_Y = f(Y)$ the corresponding graph partitions. Since $\args{\S{G}_{\ell,m}, \delta_g}$ is a geodesic space there is a midpoint $G_M$ of $G_X$ and $G_Y$. From \cite{Jain2015b}, Corollary 3.6 follows that $G_M$ is a minimum of the sample Fr\'echet function $F_2 (Z) = \delta_g(G_X, Z) + \delta_g(G_Y, Z)$. Then by \cite{Jain2015b}, Theorem 3.1 the matrix representation $\vec{G}_M$ is of the form
\[
\vec{G}_M = \frac{1}{2}\args{\vec{G}_X + \vec{G}_Y}.
\]
This implies that $G_M \in f(\S{P}_{\ell,m})$. Since $G_M$ is a midpoint and by isometry of $f$, we find that $M$ with $f(M) = G_M$ is a midpoint of $X$ and $Y$. From \cite{Lang2004}, Lemma 2.2 follows that $\S{P}_{\ell,m}$ is a geodesic space.
\end{part}


\subsection{Proof of Theorem \ref{theorem:constop}}

Standard solutions to stochastic programming problems usually assume problem formulations in Euclidean spaces. To invoke these solutions, we first send the stochastic problem \eqref{eq:stop} to an equivalent problem in the Euclidean space, then solve the problem, and finally project the solution back to the partition space. 

\subsubsection*{Step 1: Moving Problem to Euclidean Space}
To transform problem \eqref{eq:stop} to an equivalent problem in the Euclidean space, we define the pullback of function $\rho(X, Z)$ as 
\[
\rho^*: \S{X} \times \S{X} \rightarrow \R, \quad \args{\vec{X}, \vec{Z}} \mapsto \rho\Big(\pi(\vec{X}), \pi(\vec{Z})\Big),
\]
where $\pi:\S{X} \rightarrow \S{P}$ is the natural projection. The pullback $\rho^*$ of the dissimilarity $\rho$ induces pullbacks of the Fr\'echet function $F_n$ and the expectation $F_Q$. The pullback of $F_n$ is of the form
\[
F^*_n(\vec{Z}) = \frac{1}{n} \sum_{i=1}^n \rho^*\!\args{\vec{X}_{\!i}, \vec{Z}}.
\]
By $\S{F}^*_n$ we denote the set of minimizers of $F^*_n$ and by $V_n^*$ the minimum value of $F^*_n$. Similarly, 
\[
F^*_{q}(\vec{Z}) = \int_{\S{X}} \rho^*\args{\vec{X}_{\!i},\vec{Z}} dq(\vec{X})
\]
is the pullback of $F_Q$, where $q$ is a probability measure on $\S{X}$ that induces the probability measure $\S{Q}$ on $\S{P}$ as quotient measure. By $\S{F}^*_{q}$ we denote the set of minimizers of $F^*_{q}$ and by $V_{q}^*$ the minimum value of $F^*_{q}$. 

Finally, we define the distance between the mean and expected solution set by
\[
D^*\!\args{\S{F}^*_n, \S{F}^*_q} = \sup_{\vec{M}_{\!n} \in \,\S{F}^*_n} \inf_{\phantom{^X}\vec{M} \in \S{F}^*_q\phantom{^X}} \normS{\vec{M}_{\!n} - \vec{M}}{_2}.
\]

\subsubsection*{Step 2: Solving the Problem}

By the nature of stochastic programming, we assume that all samples $\S{S}_n$ are drawn i.i.d.  We will apply the following Theorem:

\medskip

\begin{framed}
\noindent
\textbf{\cite{Shapiro2009}, Theorem 5.3.}\emph{
Suppose that there exists a compact set $\S{K} \subset \R^N$ such that 
\begin{enumerate}
\item $\S{F}^*_{q} \neq \emptyset$ and $\S{F}^*_q \subseteq \S{K}$.
\item $F_q^*(\vec{Z}) < \infty$ is continuous on $\S{K}$.
\item $F_n^*(\vec{Z}) \rightarrow F_q^*(\vec{Z})$ almost surely and uniformly in $\vec{Z} \in \S{K}$.
\item $\S{F}_n^* \neq \emptyset$ and $\S{F}^*_n \subseteq \S{K}$ with probability one.
\end{enumerate}
Then 
\begin{enumerate}
\item $D^*\!\args{\S{F}^*_n, \S{F}^*_q} \rightarrow 0$ almost surely.
\item $V^*_n \rightarrow V_q^*$ almost surely.
\end{enumerate}}
\end{framed}

\medskip

We show that the assumptions of \cite{Shapiro2009}, Theorem 5.3 are satisfied. Since $\S{X}$ is compact, we can set $\S{K} = \S{X}$ and $N = \ell \cdot m$.

\setcounter{part_counter}{0}
\begin{part}
The pullback $\rho^*$ is continuous in both arguments by Lemma \ref{lemma:continuity_of_pullback}. From continuity of $\rho^*$ and compactness of $\S{X}$ follows that $F^*_{q}$ is continuous. From continuity of $F^*_{q}$ and again from compactness of $\S{X}$ follows that $F^*_{q}$ attains it minimum value. Hence, the set $\S{F}^*_{q} \subseteq \S{X}$ is non-empty and contained in a compact set. This proves the first assumption.  
\end{part}

\begin{part}
To prove the second and third assumption, we invoke the following Theorem:

\medskip

\begin{framed}
\noindent
\textbf{\cite{Shapiro2009}, Theorem 7.48.}\emph{
Let $\S{S}_n = \args{\vec{X}_1, \ldots, \vec{X}_n}$ be a sample of $n$ partitions $\vec{X}_i \in \S{X}$ drawn i.i.d. Suppose that:
\begin{enumerate}
\item For any $\vec{Z} \in \S{K}$, the function $\rho^*(\vec{X}, \cdot)$ is continuous at $\vec{Z}$ for almost every $\vec{X} \in \S{X}$. 
\item $\rho^*(\vec{X}, \cdot)$ is dominated by an integrable function for any $\vec{X} \in \S{X}$.
\end{enumerate}
Then 
\begin{enumerate}
\item $F_q^*(\vec{Z}) < \infty$ is continuous on $\S{K}$.
\item $F_n^*(\vec{Z}) \rightarrow F_q^*(\vec{Z})$ almost surely and uniformly in $\vec{Z} \in \S{K}$.
\end{enumerate}}
\end{framed}

\medskip

The assumption that the sample partitions are drawn i.i.d.~is given as stated at the beginning of the proof. The first assumption of \cite{Shapiro2009}, Theorem 7.48 is satisfied by Lemma \ref{lemma:continuity_of_pullback}. Since $\rho^*(\vec{X}, \cdot)$ is continuous on the compact set $\S{X}$ for every $\vec{X} \in \S{X}$, it attains its minimum and maximum value. Thus, there is a constant $K > 0$ such that $\abs{\rho(\vec{X}, \cdot)} \leq K$ for all $\vec{X}\in \S{X}$. This shows that $\rho^*(\vec{X}, \cdot)$ is dominated by an integrable function for any $\vec{X} \in \S{X}$. Then from \cite{Shapiro2009}, Theorem 7.48 follows the second and third assumption of \cite{Shapiro2009}, Theorem 5.3.
\end{part}

\begin{part}
Let $n \geq 1$. From continuity of $F^*_n$ and compactness of $\S{X}$ follows that $F^*_n$ attains it minimum value. Hence, the set $\S{F}^*_n \subseteq \S{X}$ is non-empty. In addition, we have $\S{F}_n^* \subseteq \S{X}$. This shows the fourth assumption.
\end{part}

\begin{part}
Since all assumptions of \cite{Shapiro2009}, Theorem 5.3 hold, both assertions hold:
\begin{enumerate}
\item $D^*\!\args{\S{F}^*_n, \S{F}^*_q} \rightarrow 0$ almost surely.
\item $V^*_n \rightarrow V_q^*$ almost surely.
\end{enumerate}
\end{part}

\begin{part}
It remains to show the Lemmata used in the proof. 

\begin{lemma}\label{lemma:continuity_of_pi}
The natural projection $\pi:\S{X} \rightarrow \S{P}$ is continuous. 
\end{lemma}

\noindent
\proof
Let $\vec{Z} \in \S{X}$. From the definition of $\delta_2$ follows
\[
\delta_2\Big(\pi(\vec{X}), \pi(\vec{Z})\Big) \leq \norm{\vec{X} - \vec{Z}} 
\]
for all $\vec{X} \in \S{X}$. Thus, $\pi$ is continuous.
\qed

\begin{lemma}\label{lemma:continuity_of_pullback}
The pullback $f^*$ of a continuous function $f:\S{P} \rightarrow \R$ is continuous. 
\end{lemma}
\end{part}

\noindent 
\proof
Since $\pi$ is continuous by Lemma \ref{lemma:continuity_of_pi} and $f$ is continuous by assumption, the pullback $f^* = f \circ \pi$ is continuous. 
\qed

\subsubsection*{Step 3: Projecting the Solution Back}

We show that 
\begin{enumerate}
\item $D^*\!\args{\S{F}^*_n, \S{F}^*_q} \rightarrow 0$ a.s. $\; \Rightarrow \; $ $D\!\args{\S{F}_n, \S{F}_q} \rightarrow 0$ a.s.
\item $V^*_n \rightarrow V_q^*$ a.s. $\; \Rightarrow \; $ $V_n \rightarrow V_q$ a.s.
\end{enumerate}
\setcounter{part_counter}{0}
\begin{part}
We have
\begin{align*}
D^*\!\args{\S{F}^*_n, \S{F}^*_q} 
&= \sup_{\vec{M}_{\!n} \in \,\S{F}^*_n} \inf_{\phantom{^X}\vec{M} \in \S{F}^*_q\phantom{^X}} \normS{\vec{M}_{\!n} - \vec{M}}{_2}\\
&\geq \sup_{\vec{M}_{\!n} \in \,\S{F}^*_n} \inf_{\phantom{^X}\vec{M} \in \S{F}^*_q\phantom{^X}} \delta_2\args{\pi\args{\vec{M}_{\!n}}, \pi\args{\vec{M}}}\\
&= \sup_{M_n \in \,\S{F}_n} \inf_{\phantom{^X}M \in \S{F}_q\phantom{^X}} \delta_2\args{M_n, M}\\
&= D\!\args{\S{F}_n, \S{F}_q}.
\end{align*}
From $D\!\args{\S{F}_n, \S{F}_q} \leq D^*\!\args{\S{F}^*_n, \S{F}^*_q}$ follows the first implication. 
\end{part}

\begin{part}
Since $F_n^*$ and $F_q^*$ are pullbacks of $F_n$ and $F_Q$, we have $V_n^* = V_n$ and $V_q^* = V_Q$. From this observation follows the second assertion. 
\end{part}

 
\subsection{Proof of Theorem \ref{theorem:CLT}}

The proof consists of the same three-step procedure as the proof of Theorem \ref{theorem:constop}. The first step can be taken from the proof of Theorem \ref{theorem:constop}.

We show the second step: Since $\S{X}$ is compact and $\rho^*$ is continuous in both arguments, we find that $\rho^*$ is Lipschitz on $\S{X}$ and p-integrable. Since $\S{S}_n$ is drawn i.i.d., we have
\[
\frac{V_n^* - V^*}{\sqrt{n}} \;\xrightarrow{\;\;d\;\;}\; \inf_{\vec{M} \in \S{F}_q*}\S{N}(0,\sigma^2(\vec{M}))
\]
by \cite{Shapiro2009}, Theorem 5.7. Finally, the back projection follows the same line of argumentation as in the proof of Theorem \ref{theorem:constop}.


\subsection{Proofs of Statements in Section \ref{sec:criteria}}
\begin{proposition}\label{prop:Cx-is-well-defined}
Let $X \in \S{P}$ be a partition. Then $\vec{X}^T\vec{X} = {\vec{X}'}^T\vec{X}'$ for all representations $\vec{X}, \vec{X}' \in X$.
\end{proposition}
\noindent
\proof
From $\vec{X}, \vec{X}' \in X$ follows that there exists a permutation matrix $\vec{P} \in \Pi$ such that $\vec{X}' = \vec{PX}$. Then we have
\[
 {\vec{X}'}^T\!\vec{X}' = (\vec{PX})^T\vec{PX} = \vec{X}^T\vec{P}^T\vec{P}\vec{X} = \vec{X}^T\vec{X}. \tag*{\qed}
\]

\medskip

\begin{proposition}\label{prop:properties-of-c_rs}
Let $\vec{C}_X = (c_{rs})$ be the compatibility matrix of partition $X \in \S{P}$. The following statements hold for all data points $z_r, z_s \in \S{Z}$:
\begin{enumerate}
\item $0 \leq c_{rs} \leq 1$.
\item If $X \in \S{P}^+$ is a hard partition, then 
\[
c_{rs} = \begin{cases}
1 & z_r \sim_X z_s\\
0 & z_r \nsim_X z_s
\end{cases}.
\]
\end{enumerate}
\end{proposition}

\noindent
\proof
Let $\vec{X} = (x_{kj})$ be a representation of $X$.
 \setcounter{part_counter}{0}
\begin{part} Since all elements $x_{kj}$ of matrix $\vec{X}$ are non-negative, we have $0 \leq c_{rs}$. From $\vec{X}^T\vec{1}_\ell = \vec{1}_m$ follows $\inner{\vec{x}_{:r},\vec{1}_\ell} = 1$ for all columns of $\vec{X}$. Then we have
\[
c_{rs} = \inner{\vec{x}_{:r},\vec{x}_{:s}} \leq \normS{\vec{x}_{:r}}{_2}\normS{\vec{x}_{:s}}{_2}.
\]
Observe that all elements $x_{kj}$ of $\vec{X}$ are from the interval $[0,1]$. Thus, we find
\[
\normS{\vec{x}_{:r}}{_2^2} = \inner{\vec{x}_{:r},\vec{x}_{:r}} \leq \inner{\vec{x}_{:r}, \vec{1}_\ell} = 1.
\]
This implies $c_{r, s} \leq 1$. 
\end{part}
\begin{part} 
The columns of $\vec{X}$ are standard basis vectors from $\R^{\ell}$, because $X$ is a hard partition. From this follows the second assertion.
\qed
\end{part}

\subsection*{Proof of Prop.~\ref{prop:properties-of-C(X,Y)}}

\setcounter{part_counter}{0}
\begin{part}
Let $\vec{C}_{\!X} = (c_{rs})$ be the compatibility matrix of hard partition $X \in \S{P}^+$. From Prop.~\ref{prop:properties-of-c_rs} follows that 
\[
c_{rs} = \begin{cases}
1 & z_r \sim_X z_s\\
0 & z_r \nsim_X z_s
\end{cases}.
\]
for all $z_r, z_s \in \S{Z}$. Suppose that $Y \in \S{P}^+$ is another hard partition with compatibility matrix $\vec{C}_Y = (d_{rs})$. Then $m_{11}$ is of the form
\begin{align*}
m_{11} &= \chi\args{\vec{C}_X, \vec{C}_Y}= \sum_{r=1}^m \sum_{s=r+1}^m c_{rs}d_{rs}.
\end{align*}
We have $c_{rs}d_{rs} = 1$ if and only if $z_r \sim_X z_s$ and  $z_r \sim_Y z_s$. This shows that $m_{11}$ counts all elements $\cbrace{z_r, z_s} \in \S{Z}^{[2]}$ that satisfy $z_r \sim_X z_s$ and  $z_r \sim_Y z_s$. Thus, $m_{11} = m_{11}^+$. For $m_{10}$, we have 
\begin{align*}
m_{10} &= \chi\args{\vec{C}_X, \vec{1}-\vec{C}_Y} = \sum_{r=1}^m \sum_{s=r+1}^m c_{rs}(1 -d_{rs}).
\end{align*}
We have $c_{rs}(1-d_{rs}) = 1$ if and only if $z_r \sim_X z_s$ and  $z_r \nsim_Y z_s$. This shows that $m_{10}$ counts all elements $\cbrace{z_r, z_s} \in \S{Z}^{[2]}$ that satisfy $z_r \sim_X z_s$ and  $z_r \nsim_Y z_s$. Thus, $m_{10} = m_{10}^+$. The relationships $m_{01} = m_{01}^+$ and $m_{00} = m_{00}^+$ follow similarly. This shows the first assertion. 
\end{part}

\begin{part} 
Let $X \in \S{P}$ be a partition. From Prop.~\ref{prop:properties-of-c_rs} follows that $0 \leq c_{rs} \leq 1$ for all elements of the compatibility matrix $\vec{C}_X = (c_{rs})$. Therefore all elements of $\vec{C}_X$ and $\vec{1}-\vec{C}_X$ are non-negative. Then the second assertion follows by definition of the function $\chi$.
\end{part}

\begin{part}
For symmetric ($m \times m$)-matrices $\vec{A}$ and $\vec{B}$, we can rewrite $\chi(\vec{A}, \vec{B})$ by 
\[
\chi(\vec{A}, \vec{B}) = \frac{1}{2}\inner{\vec{A} - \vec{D_A}, \vec{B}-\vec{D_B}},
\]
where $\vec{D_A}$ and $\vec{D_B}$ denote the diagonal matrices of $\vec{A}$ and $\vec{B}$, respectively.  For every partition $X \in \S{P}$, let $\vec{D}_{\!X}$ denote the diagonal matrix of the compatibility matrix $\vec{C}_{\!X}$. By definition, we have
\begin{align*}
\mu &= m_{11} + m_{10} + m_{01} + m_{00}\\
&= \chi\args{\vec{C}_{\!X}, \vec{C}_{\!Y}} + \; \chi\args{\vec{C}_{\!X}, \vec{1} - \vec{C}_{\!Y}} 
	+ \; \chi\args{\vec{1} - \vec{C}_{\!X}, \vec{C}_{\!Y}} + \; \chi\args{\vec{1} - \vec{C}_{\!X}, \vec{1} - \vec{C}_{\!Y}}\\
&= \frac{1}{2}\inner{\vec{C}_X-\vec{D}_X, \vec{C}_Y-\vec{D}_Y} \\
&\qquad {} + \frac{1}{2}\inner{\vec{C}_X-\vec{D}_X, \vec{1}-\vec{C}_Y-\vec{D}_{\vec{1}-\vec{C}_Y}}\\
&\qquad {} + \frac{1}{2}\inner{\vec{1}-\vec{C}_X-\vec{D}_{\vec{1}-\vec{C}_X}, \vec{C}_Y-\vec{D}_Y}\\
&\qquad {} + \frac{1}{2}\inner{\vec{1}-\vec{C}_X-\vec{D}_{\vec{1}-\vec{C}_X}, \vec{1}-\vec{C}_Y-\vec{D}_{\vec{1}-\vec{C}_Y}}.
\end{align*}
We set $\vec{E}_X = \vec{C}_X - \vec{D}_X$ and $\vec{E}_Y = \vec{C}_Y - \vec{D}_Y$. Observe that $\vec{D}_{\vec{1}-\vec{A}} = \vec{I} - \vec{D_A}$ for every ($m\times m$)-matrix $\vec{A}$, where $\vec{I}$ denotes the identity matrix. Then 
\begin{align*}
2\mu &= \inner{\vec{E}_X, \vec{E}_Y} + \inner{\vec{E}_X, \vec{1}-\vec{I} - \vec{E}_Y} + \inner{ \vec{1}-\vec{I} - \vec{E}_X, \vec{E}_Y} + \inner{ \vec{1}-\vec{I} - \vec{E}_X, \vec{1}-\vec{I} - \vec{E}_Y}\\
&= \inner{\vec{E}_X, \vec{E}_Y} + \inner{\vec{E}_X, \vec{F} - \vec{E}_Y} + \inner{ \vec{F} - \vec{E}_X, \vec{E}_Y} + \inner{ \vec{F} - \vec{E}_X, \vec{F} - \vec{E}_Y},
\end{align*}
where $\vec{F} = \vec{1}-\vec{I}$. Expanding and summarizing gives
\begin{align*}
2\mu &= \inner{\vec{E}_X, \vec{E}_Y}  +  \inner{\vec{E}_X, \vec{F}}  - \inner{\vec{E}_X, \vec{E}_Y} + \inner{\vec{F}, \vec{E}_Y}  - \inner{\vec{E}_X, \vec{E}_Y} \\
&\qquad {} + \inner{\vec{F}, \vec{F}} - \inner{\vec{F}, \vec{E}_Y} - \inner{\vec{F}, \vec{E}_X} +  \inner{\vec{E}_X, \vec{E}_Y} \\
&= \inner{\vec{F}, \vec{F}}.
\end{align*}
Note that $\normS{\vec{1}}{_2^2}  = m^2$, $\normS{\vec{I}}{_2^2}  = m$ and $\inner{\vec{1}, \vec{I}} = \inner{\vec{I}, \vec{I}} = m$. Thus,
\begin{align*}
\inner{\vec{F}, \vec{F}} &= \inner{\vec{1}-\vec{I}, \vec{1}-\vec{I}} 
= \normS{\vec{1}}{_2^2} - 2\inner{\vec{1}, \vec{I}} + \normS{\vec{I}}{_2^2}\\
&= m^2 -2m + m = m (m-1).
\end{align*}
This shows $\mu = m(m-1)/2 = N$. 
\end{part}
\qed

\subsection*{Proof of Prop.~\ref{prop:continutity_1-5}}
The function $\chi$ as an inner product of strictly upper triangular matrices is continuous in both arguments. The composition of continuous functions is continuous. Critical points are points where the denominator of $\rho_1 - \rho_5$ becomes zero. These points form a set of Lebesgue measure zero. Continuity of $\rho_3$ follows from the assumption that $m > 1$.

\subsection*{Proof of Prop.~\ref{prop:properties-of-xyz}}
\setcounter{part_counter}{0}
\begin{part}
Suppose that $X, Y \in \S{P}^+$ are hard partitions with representations $\vec{X}$ and $\vec{Y}$, respectively. The rows $\vec{x}_{p:}$ of $\vec{X}$ are binary vectors satisfying
\[
x_p^+ = \abs{\S{C}_p(X)} = \sum_{j=1}^m x_{pj} = \inner{\vec{x}_{p:}, \vec{1}_m} = x_p.
\] 
Since $y_q^+$ and $y_q$ only differ from $x_p^+$ and $x_p$ in notation, we also have $y_q^+ = y_q$. Finally, we have
\[
z_{pq}^+ = \abs{\S{C}_p(X) \cap \S{C}_q(Y)} = \sum_{j=1}^m x_{pj}\cdot y_{qj} = \inner{\vec{x}_{p:}, \vec{y}_{q:}} = z_{pq}.
\]
\end{part}

\begin{part}
The second assertion follows from the fact that representation matrices have non-negative elements only. 
\end{part}

\begin{part}
We have
\begin{align*}
\sum_{p=1}^\ell x_{p} = \sum_{p=1}^\ell \inner{\vec{x}_{p:}, \vec{1}_m} = \sum_{p=1}^\ell\sum_{j=1}^m x_{pj} = m,
\end{align*}
where the last equality follows from $\vec{X}^T\vec{1}_\ell = \vec{1}_m$. Obviously, the same holds for $\sum_q y_q$.
\end{part}
\qed

 \end{appendix}

\end{document}